\documentclass{article}
\usepackage{spconf,amsmath,graphicx}
\usepackage{subfigure}

\title{Forward Vehicle Collision Warning Based on Quick Camera Calibration}
\name{Yuwei Lu$^{1}$ \qquad Yuan Yuan$^{1}$ \qquad Qi Wang$^{1,2*}$ \thanks{$^{*}$Corresponding author. This work was supported by the National Key Research and Development Program of China under Grant 2017YFB1002202, National Natural Science Foundation of China under Grant 61773316 and 61379094, Fundamental Research Funds for the Central Universities under Grant 3102017AX010, and the Open Research Fund of Key Laboratory of Spectral Imaging Technology，Chinese Academy of Sciences. \protect \\ \copyright 2018 IEEE. Personal use of this material is permitted. Permission from IEEE must be obtained for all other uses, in any current or future media, including reprinting/republishing this material for advertising or promotional purposes, creating new collective works, for resale or redistribution to servers or lists, or reuse of any copyrighted component of this work in other works. }}
\address{$^1$School of Computer Science and Center for OPTical IMagery Analysis and Learning (OPTIMAL), \\
	Northwestern Polytechnical University, Xi'an 710072, Shaanxi, P. R. China\\
	$^2$Unmanned System Research Institute (USRI),\\ Northwestern Polytechnical University, Xi'an 710072, Shaanxi, P. R. China\\}

\begin{document}
%
\maketitle
\begin{abstract}
Forward Vehicle Collision Warning (FCW) is one of the most important functions for autonomous vehicles. In this procedure, vehicle detection and distance measurement are  core components, requiring accurate localization and estimation.  In this paper, we propose a simple but efficient forward vehicle collision warning framework by aggregating monocular distance measurement and precise vehicle detection. In order to obtain forward vehicle distance, a quick camera calibration method  which only needs three physical points to calibrate related camera parameters is utilized. As for the forward vehicle detection, a multi-scale detection algorithm that regards the result of calibration as distance priori is proposed to improve the precision. Intensive experiments are conducted in our established real scene dataset and the results have demonstrated the effectiveness of the proposed framework.
\end{abstract}
\section{Introduction}
\label{sec:intro}
Significant effort has been made on the safety of road  vehicles in recent decades. Over 10 million people are injured yearly worldwide in road accidents. Automatic driving and assistance driving are  the research directions to increase the safety of passengers and of vehicles. Forward vehicle collision warning is  the fundamental  function in both automatic driving and assistance driving.

A range of devices mounted on the vehicle could provide the solution to  this problem \cite{Widmann1998Development,YUAN2018202}. The traditional systems available today are typically based on radar sensors \cite{typical}.  However, the narrow field of view and the poor lateral resolution limit the performance of these systems. From a technological point of view, fusion of radar and vision information is an attractive approach. In such  a system \cite{fusion,WangYLW16} the radar provides accurate distance and  velocity, while vision gets exact locations of  forward vehicles. However, this solution is expensive and complex in practical applications.  

Considering these practical difficulties, a simple but efficient forward vehicle collision warning framework is proposed using only vision information in this paper. The proposed framework includes two stages. The camera calibration stage obtains distance from forward vehicles to  the camera,  while the vehicle detection stage based on the distance  gets exact locations of  forward vehicles.  The main contributions of our framework are as follows: First, a simple but effective framework is proposed  for forward vehicle collision warning. Since it is based on vision information, the framework is inexpensive. Second, distance information is utilized  to improve the performance of detecting forward vehicles.

The rest of this paper is organized as follows. Section \ref{sec:rel} introduces the related work and Section \ref{sec:met} describes the proposed framework. Experimental results are demonstrated in Section \ref{sec:exp} while   conclusion is presented in Section \ref{sec::con}.




\section{Related Work}
\label{sec:rel}

\subsection{Camera Calibration} \label{subsec::calib}
Camera calibration has been studied extensively in computer vision and photogrammetry. According to the dimension of the calibration object, calibration methods can be roughly classified into two categories as follows:

\textbf{object-based calibration}. Techniques in this category are required to observe a calibration object \cite{3dcalibration,Tsai1987A,Zhang2000A,SturmM99,Zhang2004Camera}. Some of these methods \cite{3dcalibration,Tsai1987A} require   the geometry information of object in 3D space with very good precision.  These approaches always need an expensive calibration apparatus, and a complex setup.  Therefore, some methods calibrate camera by observing a planar pattern shown at a few different orientations \cite{Zhang2000A,SturmM99}. Since such a  calibration pattern is easy to be made, the setup becomes easier. In order to make calibration easier, Zhang \cite{Zhang2004Camera} proposes one-dimensional object based calibration.  It uses less knowledge of the observation compared to 2D and 3D object-based calibration methods.

\textbf{Self-calibration}. Techniques in this category don't utilize any calibration objects \cite{Pollefeys2000Multiple}. By moving a camera in a static scene, the internal parameters of the camera will be estimated with image information alone. Though no calibration objects are necessary, a large number of parameters still need to be estimated. Computational complexity will be greatly increased.

\subsection{Vehicle Detection}
Considering the practical applications, deep learning methods will not be reviewed in this section. Vehicle detection approaches  are mainly divided into two types : template-based and appearance-based.

\textbf{Template-based methods}. Methods in this category apply predefined patterns from	the vehicle class and perform correlation between the image	and the template. Li \textit{et al.} \cite{eccvLiWZ14} propose an And-Or model	that integrates context and occlusion for  detecting vehicles. Felzenszwalb \textit{et al.} \cite{FelzenszwalbGMR10} propose deformable part models to structure template model. Each model is composed  of  parts with different viewpoints. They detect vehicles by integrating various parts of vehicles.   Wang \textit{et al.} \cite{Wang2016Probabilistic}  also propose	a probabilistic inference framework based on part models for	improving detection performance.  Since these methods detect vehicles by matching template, they are time consuming.

\textbf{Appearance-based methods}. Appearance-based methods learn the features of  vehicles from a set of training images which should capture the	variability in vehicle appearance. Usually, appearance models	treat a two-class pattern classification problem: vehicle and	nonvehicle. Zheng \textit{et al.} \cite{ZhengL09} design image strip features based on the vehicle structure for vehicle detection. Since 	features come from the side view of the vehicle, this detector	is sensitive to the viewpoint.    Dollar \textit{et al.} \cite{Dollaracf14} propose aggregate channel features (ACF) and Yuan \textit{et al.} \cite{pure} improve the features for detection.  ACF utilizes color information of the objects to improve the performance.


\section{Our Method}
\label{sec:met}
As mentioned before, the proposed framework includes two stages: camera calibration and vehicle detection. To simplify the calibration course,  a point-based calibration method \cite{Chenchen2014A} is employed to get camera parameters and to calculate distance from the forward car.  During the  detection course, we expand original ACF detector \cite{Dollaracf14} into a distance-based multiple scale detector.     The distance is not only used for forward collision warning, but also employed for improving vehicle detection.

%
%

\begin{figure}[htb] 
	
	
	\begin{minipage}[b]{0.48\linewidth}
		\centering
		\centerline{\includegraphics[width=4.0cm ]{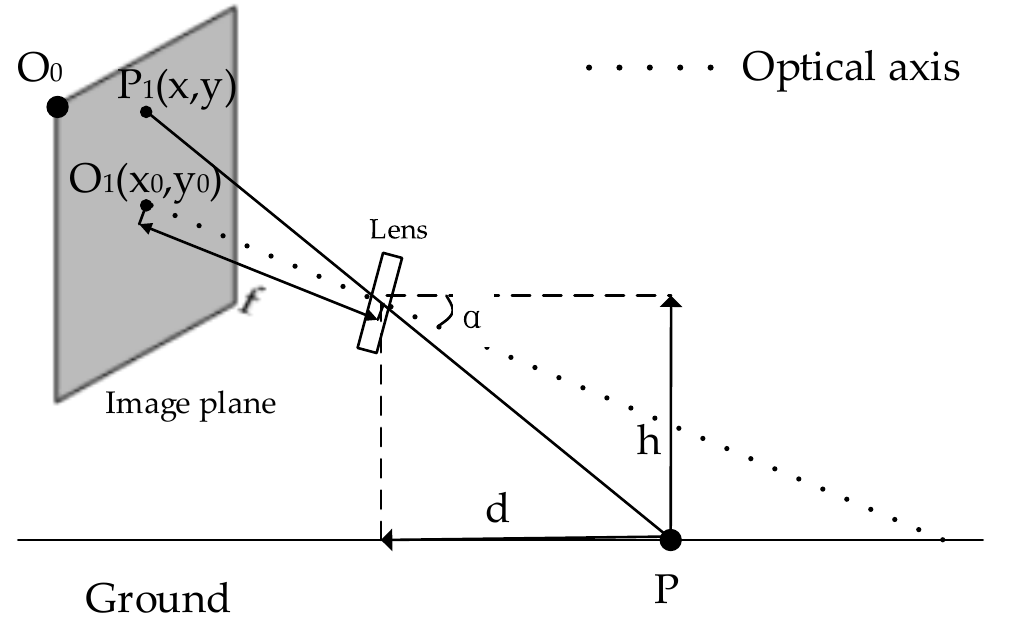}}
		\centerline{(a) }\medskip
	\end{minipage}
	\hfill
	\begin{minipage}[b]{0.48\linewidth}
		\centering
		\centerline{\includegraphics[width=4.0cm]{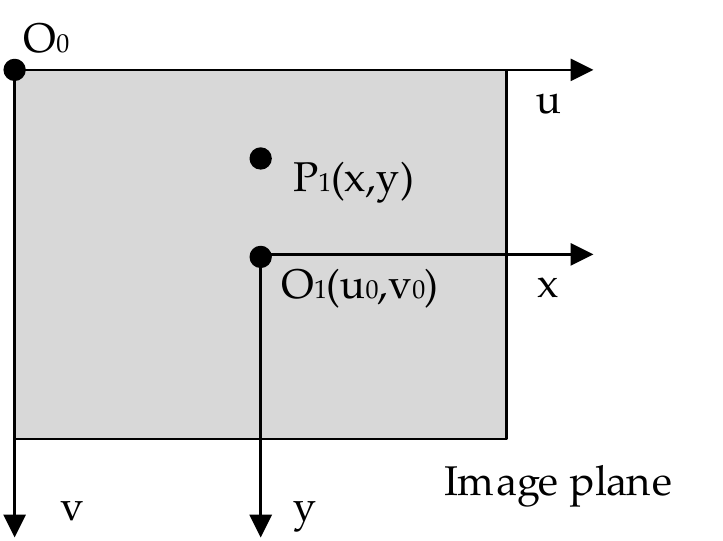}}
		\centerline{(b)}\medskip
	\end{minipage}	
	\caption{The pinhole imaging model of forward point P. (a) is the projection model and Eq \ref{eq.1} is derived from it; (b) shows the relation of idealized image coordinate system $xO_1y$ to camera’s pixel location coordinate system $uO_0v$. }\label{figcalib}
\end{figure}

\begin{figure*} [htb] 
	\centering
	\centerline{\includegraphics{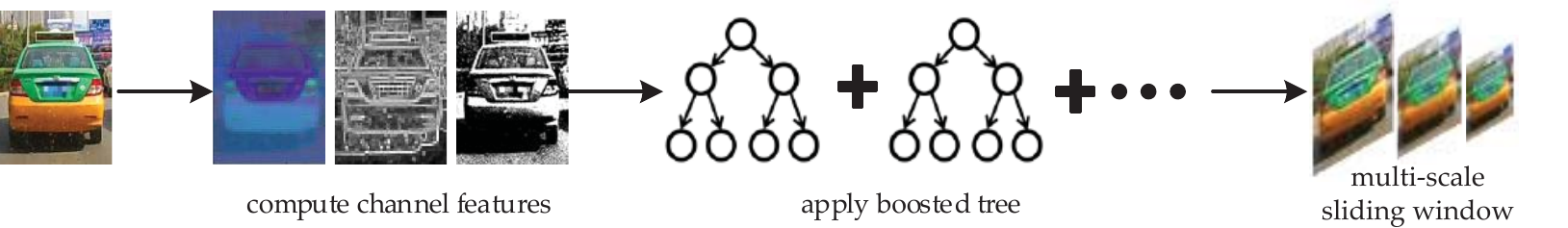}}
	\caption{Overview of our detection framework. }\label{figdet}
\end{figure*}

\subsection{Point-based calibration} \label{subsec::calibrate}

Camera calibration is  a necessary step in distance measuring with monocular vision. In engineering practice, the object distances are usually considerably larger than the focal length of camera. Hence, the pinhole camera model can be used to measure the distance. The geometry relationship of actual point $P$ on the ground and its projection point on the image plane $P_1$ is shown in Fig \ref{figcalib}(a).  According to \cite{Wang2001Study}, the distance from point $P$ to camera is:
\begin{equation} \label{eq.1}
	d = \dfrac{h}{\tan(\alpha + \arctan[|(y_0-y)/f|])},
\end{equation}
Here, $\alpha$ is the pitch angle of the camera; $h$ is the height of the camera from the ground; $(x_0,y_0)$ is the cross point of optical axis of the camera and the image plane; and $y$ is the vertical coordinate of $P_1$. In order to simplify the calibration process, let $dx$, $dy$ denote the physical dimension of one pixel along the x-axis and the y-axis separately. Then the coordinates of point $P_1$ in the image physical coordinate plane $xO_1y$ and its position in the image pixel reference frame $uO_0v$ are related by the transformation equation:

\begin{equation} \label{eq.2}
	u = \frac{x}{dx}+u_0, v = \frac{y}{dy}+v_0,
\end{equation}
In theory, as the corresponding pixel location of $(x_0,y_0)$, $(u_0,v_0)$ usually locates in the center of image. But in fact, there might be slight departure due to fabrication. In that case, $u_0$ and $v_0$  need to be measured.     So, Eq. \ref{eq.1} can be expressed as
\begin{equation} \label{eq.3}
	d = \dfrac{h}{\tan(\alpha + \arctan[|(v_0-v)/f_y|])}.
\end{equation}
Here, $f_y=f/dy$. Hence, we can get the distance $d$ by solving the ratio $f_y$ rather than calculating the optical length and pixel physical dimension separately. In practice, the height of camera $h$ can be measured  after the camera is mounted on the car. Only $f_y,v_0,\alpha$ need to be calibrated.  Based on Eq. \ref{eq.3}, three fixed points are utilized to estimate camera parameters in \cite{Chenchen2014A}. With the input of height $h$, distances information of three calibration points $(d_1,d_2,d_3)$ and corresponding coordinates $(u_1,v_1),(u_2,v_2),(u_3,v_3)$, $f_y,v_0,\alpha$ will be solved.

This algorithm needs only three fixed points to complete the calibration,  which greatly reduces the computational complexity.    Different from traditional ones in Section \ref{subsec::calib},  this  method  estimates less parameters (only $f_y,v_0,\alpha$ and $h$) with the purpose of measuring distance. Estimating less parameters  makes  the calibration course easy to setup.

\subsection{Multi-scale detection} \label{ourdetetion}

When detecting forward vehicles, one of the greatest challenge is that vehicles have various scales at different distances. Multi-scale and multi-aspect ratio make this problem difficult.  Due to perspective principle of  the camera, the features  of vehicles will change with different size. The structural feature is significant when the forward vehicle is near. However,  when forward vehicles are far, they are made up of a few pixels in the image plane. We can hardly get structural features in this situation.  Therefore, we apply color based features \cite{Dollaracf14} to detect forward vehicles and  distance information is employed to tackle multi-scale problem.

\begin{figure}[htb] 
	
%
%
	
		\begin{minipage}[b]{0.48\linewidth}
		\label{figcalib-a}
			\centering
			\centerline{\includegraphics[width=4.0cm]{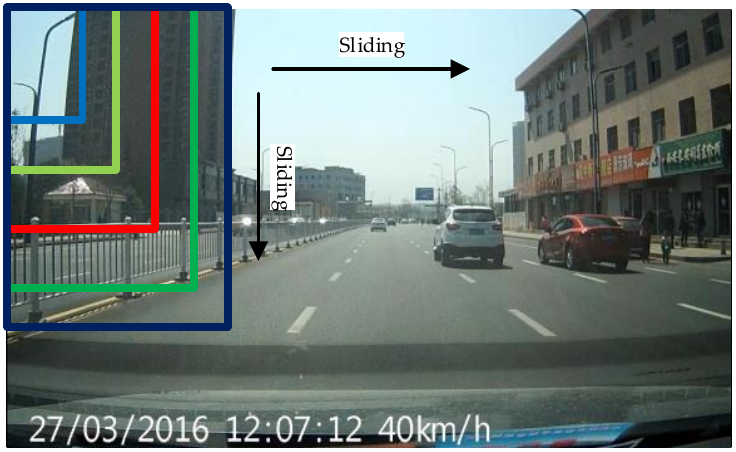}}
			\centerline{(a) original ACF detector}\medskip
		\end{minipage}
		\hfill
		\begin{minipage}[b]{0.48\linewidth}
			\centering
			\centerline{\includegraphics[width=4.0cm]{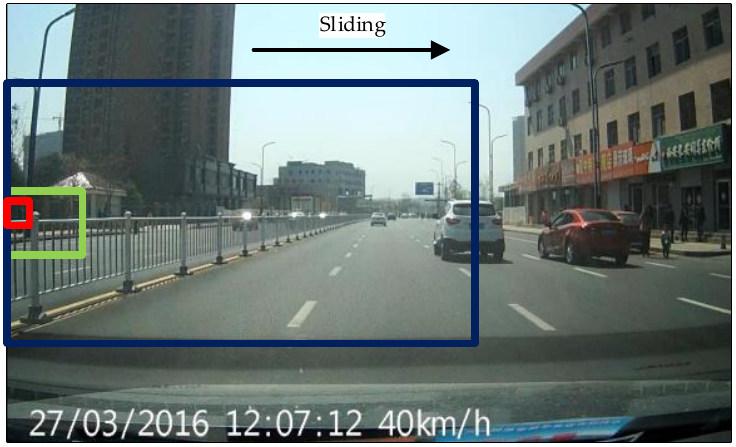}}
			\centerline{(b) the proposed detector}\medskip
		\end{minipage}	
	\caption{The difference between original ACF detector and our detector.}\label{sliding}
\end{figure}

The proposed detection framework is exhibited in Fig. \ref{figdet}. Given an input image $I$, we compute its channel features. Then the boosting is used to train and combine decision trees over these channel features to  distinguish object from background. Next, a distance based multi-scale sliding window approach is employed to detect vehicles.  Fig.\ref{sliding} illuminates the major differences between original ACF detector and the proposed detector. Windows with various scales will slide the whole image in the ACF detector, while the proposed detector uses several windows with certain scale and aspect ratio to slide part of the image.  Due to applying diverse windows in different vertical coordinates, the proposed method will be less time consuming.

The scale of sliding windows is related to distances between cars and the camera. Eq. \ref{eq.3} can be changed into the following form:
\begin{equation} \label{eq.13}
v = v_0 - f_y \tan(\arctan\frac{h}{d}- \alpha).
\end{equation}
Eq. \ref{eq.13}  is the foundation of multi-scale detection with distance prior. It  indicates that if $v_0, f_y$ and $\alpha$ are estimated, the vertical coordinate can be obtained by giving the real distance $d$.       Therefore, we build a mapping from forward distance to the vertical coordinate in the image plane.

The distance prior can be calculated with a calibrated camera according to Eq. \ref{eq.13}. When camera calibration  is completed, we can not only obtain distance from for ward vehicles to the camera, but also get locations in the image plane  according to the distance conversely. Table \ref{scale} demonstrates some scales of sliding windows in different distance.

\begin{table}[!hbp] 
	\begin{center}		
		\caption{The change of aspect ratio at different distances}
		\begin{tabular}{c| c | c} 
			\hline
			{\bf distance / m} & {\bf  scale}  & {\bf vertical coordinate / pixel} \\ \hline
			5 & 400 $\times$ 275  & 482 \\
			10  & 110 $\times$ 95  & 325 \\
			20 &  50 $\times$ 45 & 260 \\			
			\hline
		\end{tabular}	 \label{scale}	
	\end{center}
\end{table}

On account of the mapping from forward distance to the vertical coordinate, we don't need to slide various size of windows in the whole image. According to the distance prior, we can use multiple scale sliding windows in different vertical coordinates on the image.  The size of sliding window can be determined by statistics. During the statistical process of window size,  we discover that not only the scale but also the aspect ratio of forward vehicles will change as the distance varies. When the vehicle is far,  its scale is small and the aspect ratio will be approximate to 1:1. However, the aspect ratio of vehicles will change into nearly 1.5:1 when they are close to the camera, e.g. 5 meters.  

The reason for the change of  aspect ratio  is  the extension distortion caused by wide-angle camera. Since drive recorders always utilize the wide-angle camera, the change of aspect ratio does exist in practice.  Although  extension distortion can be calibrated and corrected, we don't calibrate it in practice.  Calibrating more parameters will make calibration course more difficult to setup. However, when measuring the distance of a forward vehicle,  extension distortion will not affect this course. For the reasons above,  extension distortion is ignored in the calibration course.



According to the distance prior, our multi-scale detection  could  search vehicles in different distance with a certain scale. The main advantages of our multi-scale detection are as follows: First, we relieve the multi-scale problem in forward vehicle detection. Then, due to sliding window with a certain scale in different locations in an image, the proposed method speeds up the detection course. Hence,  our multi-scale detection can be faster than the original one and reach 50fps on CPU.

\section{Experiment}
\label{sec:exp}

To demonstrate the capabilities of the presented approach, extensive experiments are conducted and evaluated in an Intel i5 quad core CPU with 3.20GHz.   In this section, we will introduce the experiment from the following two aspects: calibration and detection.


\begin{table}[!hbp] 
	\begin{center}		
		\caption{Experimental results of camera calibration}
		\begin{tabular}{c| c | c | c | c } 
			\hline
			{ Car No.}   & { d / m} & {d' / m} & $e*$ / m & $e_r$ / \% \\ \hline
			1 &  5.00   & 5.00  & 0.00 &  0.00 \\
			2 &   7.00  & 6.98 & 0.02 &   0.29\\
			3 &  9.00   & 9.08  & 0.08 &  0.89 \\	
			4 &   11.00  & 11.11  & 0.11 &   1.00\\	
			5 &   15.00  & 15.26 & 0.26 &   1.73\\	
			6 &   17.00  & 17.31 & 0.31 &   1.82\\			
			\hline
		\end{tabular}	\label{test}	
	\end{center}
\end{table}

\subsection{Validation of calibration method}

In our experiment, the images comes from the camera of ordinary   driving recorder and its size is $1280 \times 720 $.   The height of the camera is 122.5cm, and three fixed points used for the calibration are 4m, 5m and 7m away from the camera. Their vertical coordinate are 461, 428, 383.   Following the calibration steps mentioned in Section \ref{subsec::calibrate}, we obtain camera parameters for measuring distance.     The calibration results are $\alpha=0.1194rad, f_y=1094.313$ and $v_0 = 363.331$. Then a set of test cars are substituted into the algorithm to detect its measurement error. The estimated distance is denoted by $d'$.  The absolute error and relative error can be expressed separately as $e^*=|d-d'|$ and $e_r=e^*/d$. The measuring results are demonstrated in Table \ref{test}.

As illustrated in Table \ref{test}, this algorithm performs well when the points are near, and relative errors increase with the distance becomes far.   This is because that along with the object getting farther, one pixel on the image covers longer distance. In other words, if one pixel represents several centimeters in the near, it may represent several meters in the far distance. It is an inherent defect of monocular vision.

\subsection{Validation of distance based detection}
In order to illustrate the performance of the proposed detection method, comparisons are made between our detector and \cite{DPMpami,eccvLiWZ14,ohnbar14,Dollaracf14}. All of these methods are trained by KITTI car detection dataset \cite{kitti} and tested on 5400 images of real scene collected by ourselves. The test images come from 30 different driving videos taken by the same recorder. These videos cover urban road, highway, night, rainy and other situations. Each video is 3 minutes with 30 fps, and test images are selected every one second.   Considering the limitation of computing resource in the practical application, deep learning methods will not be compared in this section. 


Table \ref{tab4} shows the comparisons of  detection rate and FPPI (false positive per image). Benefiting from distance prior, our detector has the knowledge of vehicle size in different vertical coordinates. FPPI decreases obviously, which means  less false detection occurs during our framework. Because we have certain scales in different vertical coordinates, our detector performs better. Besides, certain scales also decrease the number of sliding windows. It  also makes the proposed detector faster than others. 

Our multi-scale detection framework can achieve 50 frames per second on CPU, which can  meet the requirements of other  automatic driving and assistance driving  applications besides forward vehicle collision warning in the future.



\begin{table}[!hbp] 
	\begin{center}
		\caption{	Comparison of various detection methods}
		\begin{tabular}{c| c | c |c } 
			\hline
			{ } & {  Detection rate} & {  FPPI }  & { 	Time(s)/frame }  \\ \hline
			 DPM \cite{DPMpami} & 91.23\%  &  0.098  & 4.0   \\
		 And-Or \cite{eccvLiWZ14} & 89.08 \% & 0.133 & 3.0  \\
		SubCat	\cite{ohnbar14} & 92.70 \% & 0.087 & 0.7  \\
		ACF	\cite{Dollaracf14} & 94.02 \% & 0.065 & 0.04  \\
			Ours & 96.61\% & 0.046 & 0.02  \\	
			\hline
		\end{tabular} \label{tab4}
	\end{center}
\end{table}



\section{Conclusion}
\label{sec::con}


In conclusion,  we propose a simple but efficient forward vehicle collision warning framework by aggregating monocular distance measurement and precise vehicle detection.  Point based calibrating algorithm  greatly reduces the computational complexity and can be easily  achieved. Multi-scale detection has taken excellent advantage of distance prior to  improves accuracy and decrease time consumption. The processing speed of the proposed framework  can achieve 50 fps on CPU  and the experiments have exhibited its outstanding performance.


\bibliographystyle{IEEEbib}
\bibliography{strings,refs}

\end{document}